# Representation Robustness under Executable Reasoning Constraints in Large Language Models for Mathematical Problem Solving

Sagnik Nath[1[0000-0002-5225-952X]], Edith Aurora Graf[2[0000-0002-7650-477X]], Liang Zhang[3[0009-0002-0017-2569]] and Diego Zapata-Rivera[4[0000-0002-0620-7622]]

[1] University of California, Santa Cruz, Santa Cruz CA 95060, USA
sanath@ucsc.edu
[2] Independent Researcher
eag2718@gmail.com
[3] Medical School, University of Michigan–Ann Arbor, Ann Arbor MI 48109, USA
zhlian@umich.edu
[4] ETS Research Institute, Princeton, NJ 08541, USA
dzapata@ets.org

Corresponding author: Sagnik Nath

**Abstract.** Large language models (LLMs) are increasingly evaluated on mathematical problem solving yet prior work often treats representationally equivalent problem formulations as interchangeable and conflates reasoning errors with interface failures. In this paper, we investigate representation robustness in LLM-based mathematical problem solving by systematically varying surface representations (variants) of the same underlying problems, including story problems, word-equations, symbolic equations, and isomorphic paraphrases. Using a curated dataset of mathematically equivalent problems, we evaluate 5 contemporary LLMs under a direct answer generation interaction condition. In this baseline setting, we observe substantial representational sensitivity such as models frequently changing correctness across equivalent formulations, with nontrivial flip rates across story, symbolic, and word-equation variants. We further find systematic regressions under isomorphic reformulations, showing that even subtle paraphrase level changes can degrade performance even when mathematical structure is preserved. We then evaluate a second interaction condition (code-augmented) that constrains models to externalize reasoning as executable Python code, executed locally for validation. While this executable interface reveals strong latent reasoning capability in models that performed poorly under direct prompting, it does not uniformly improve robustness. Instead, failures are redistributed across interaction layers, shifting from opaque reasoning errors to protocol violations and execution failures. Even when executable reasoning succeeds, representation sensitivity often persists. Together, these results show that reasoning scaffolds do not eliminate representational brittleness but expose new tradeoffs between correctness, reliability, latency, and cost. We argue that representation should be treated as a first-class interface design variable in LLM evaluation and deployment with implications for the design of AI-assisted problem solving systems.

# 1 Introduction

Large language models (LLMs) continue to demonstrate impressive fluency in mathematical problem solving, often producing correct answers and coherent explanations for grade school and algebraic tasks [1]. As a result, benchmark-driven evaluations of math reasoning in LLMs have become increasingly common, with performance improvements frequently interpreted as evidence of advancing reasoning capability. However, a growing body of work has begun to question whether these evaluations reliably capture genuine reasoning or instead reflect sensitivity to surface form, memorized patterns, and interaction artifacts rather than an actual problem understanding( [2], [3]).

A recent critique of popular math benchmarks is that LLM performance can vary substantially across different representations of the same underlying problem, even when only numerical values or superficial phrasing are changed( [3], [4] , [5]). These findings suggest that single point accuracy metrics may obscure significant variance and fragility in model behavior and that apparent reasoning success may not generalize across equivalent representations [6]. While such work has been instrumental in exposing limitations of benchmark centric evaluation, it has largely framed the issue as a problem of dataset design or model generalization [2].

In this paper, we take a complementary HCI perspective: ***we treat mathematical problem representations as interface design choices that structure how humans and LLMs interact.*** From this viewpoint, story problems, word-equations, symbolic equations, and isomorphic paraphrases are no longer interchangeable inputs but distinct representational interfaces that output different interpretive and computational behaviors in LLMs. Decades of research in education and cognitive science have shown that humans exhibit systematic performance differences across mathematically equivalent representations depending on how problems are framed and externalized( [7], [8]). Yet, despite claims that LLMs can be driven to reason symbolically ( [9], [10]) , little work has examined whether their behavior exhibits comparable representational robustness (or comparable fragility) under controlled interaction conditions (e.g. [11], [12]) .

Guided by this perspective, we investigate the following research questions:

- RQ1: How robust are LLMs to representationally equivalent formulations of the same mathematical problem?
- RQ2: How does externalizing reasoning into executable code affect representational robustness and failure behavior?
- RQ3: When executable reasoning fails, how are errors distributed across interaction layers such as protocol adherence, execution, and mathematical correctness?

- RQ4: What tradeoffs in accuracy, reliability, latency, and cost arise when reasoning is mediated through executable interfaces?

To study these RQs, we evaluate contemporary LLMs on a curated set of mathematically equivalent problems expressed across multiple representations, including contextual word problems, symbolic forms, word-equation paraphrases, and isomorphic variants with altered numerical structure. We compare two interaction conditions: a ***baseline condition*** in which models produce answers directly in natural language, and a constrained ***code-augmented*** executable-reasoning condition that requires models to externalize reasoning as Python code which is then executed client-side. This design allows us to disentangle failures of mathematical reasoning from failures of interaction contracts and tool use.

The remainder of the paper is organized as follows. Section 2 describes related work in educational research, LLM evaluation, and tool augmented reasoning. Section 3 describes the dataset construction and experimental methodology. Section 4 presents empirical results across models, representations and interaction conditions. Section 5 discusses implications for the design and evaluation of AI assisted math problem solving systems. Section 6 concludes with future work.

# 2 Background and Related Work

## 2.1 Representation and Problem Solving in Human Cognition

Research in education and cognitive psychology has long established that the representation of a problem plays a critical role in how humans reason, plan and make errors. For example, [8] showed that students often succeed or fail not because of differences in underlying mathematical structure but because *particular representations* differentially engage comprehension demands and solution strategies. Their work emphasized that problem representations shape how a solver engages with a task, leading to predictable error patterns even when semantic equivalence is preserved. Studies of isomorphic algebra story problems [7] similarly demonstrated that representational form and problem templates strongly influence solution behavior and transfer despite identical underlying equations. Such findings establish representation as a central explanatory factor in problem solving behavior rather than a peripheral design detail in designing math solving LLM models.

## 2.2 Representation Sensitivity and Tool Augmented Reasoning in LLMs

Recent evaluations of math reasoning in LLMs have questioned the reliability of benchmark-based accuracy metrics [6]. Studies using controlled variants of the same underlying math problems (e.g. varying numerical values, phrasing, or structure) show that LLM performance can differ noticeably across representationally equivalent representations. This variability challenges the assumption that high benchmark scores alone reflect stable reasoning, indicating that the apparent success may depend on surface level cues rather than invariant mathematical structure [13].

A parallel line of work has proposed tool augmented reasoning as a means of improving LLM reliability. Approaches such as Chain of Thought (COT) prompting [14], Program Aided Language (PAL) models [11], and tool use frameworks [12] encourage models to externalize intermediate reasoning steps, often by generating executable code or structured representations that can be verified. Overall, these methods have been shown to reduce arithmetic errors and improve accuracy on certain classes of mathematical tasks.

However, from an HCI perspective, such tool augmented reasoning approaches can be understood not merely as a performance enhancement but as an interaction design intervention. Requiring models to reason through code or structured artifacts introduces new constraints, affordances and failure modes that reshape how reasoning is expressed and validated. Importantly, most prior work in this space evaluates tool augmented reasoning under a ***fixed problem*** representation, implicitly assuming that representational form does not significantly interact with the reasoning interface itself.

***As a result, existing research leaves open a key question as to whether external reasoning scaffolding mitigates representational sensitivity or whether it simply redistributes errors across tool use interaction layers such as protocol compliance, execution correctness and mathematical validity***. Addressing this gap requires examining representational form and reasoning scaffolds together rather than in isolation. The current work is motivated by this need and treats both representation and tool mediation as first-class design variables in the evaluation of LLM as tutors in math problem solving.

# 3 Method

## 3.1 Study Framing and Experimental Goals

This study examines how problem representation and interaction modality jointly shape LLM behavior in mathematical problem solving. Rather than treating representation as a surface level input choice, we treat it as a first-class interaction variable. Similarly, we treat executable (code-augmented) reasoning not as an internal optimization technique but as an alternative interaction interface with its own constraints, affordances, and failure modes.

The experiment follows a fully crossed design over problem representation, interaction condition, and model. The unit of analysis is a single model API call for a given problem, its representation variants and interaction condition. Dependent measures include correctness, failure type, latency and estimated API cost.

## 3.2 Dataset Construction and Representation Design

The dataset is derived from 85-word problems [15] originally presented as "story problems" in the appendix of [7]. These problems were explicitly designed to study representational equivalence and isomorphism in human problem solving. The math problems in this dataset span classic middle school through early high-school algebra topics (e.g. linear equations, rates and speed problems, work problems, mixtures, percentages, ratios, simple interest, inverse variation, basic geometry and introductory arithmetic/geometric progressions) designed to be solvable using one or two variable algebraic reasoning typical of grades 6–10 before formal calculus or advanced algebra is introduced. We adopt these problems not for their difficulty or novelty but mainly because they provide controlled families of mathematically identical problems that allow for multiple equivalent representations.

Each problem is instantiated in multiple representational forms, or variants as was established in [8].

- **Story** problem, or the original natural language narrative ("*George rode out of town on the bus at an average speed of 20 miles per hour and walked back at an average speed of 3 miles per hour. How far did he go if the entire trip took six hours?*")
- **Symbolic** equation form, or the formal mathematical notation ("*Solve for x: x (1/20 + 1/3) =6*")
- **Word-equation** paraphrase, or the textual description of algebraic relationships ("*Starting with some number, if I multiply it by the sum of 1 divided by 20 and 1 divided by 3, I get 6. What number did I start with?*")

For baseline experiments, we also include **isomorphic** paraphrases derived from the original story problem. The key feature of isomorphs is that they share the same underlying mathematical structure, but the language can be different( [16], [7], [17]). In our case, the natural language of the story is also retained except some of the numerical digits are altered, specifically ***surface level isomorphism*** as defined in [18] ("*George rode out of town on the bus at an average speed of 10 miles per hour and walked back at an average speed of 2 miles per hour. How far did he go if the entire trip took five hours?*"). This allows us to test whether model performance reflects invariant mathematical reasoning or instead depends on sensitivity to specific surface features such as numbers, magnitudes or lexical patterns. This parallels findings from human problem solving studies (e.g. [18], [19]) which show that performance varies systematically across isomorphic variants despite identical underlying mathematical structure. It also potentially mitigates the risk of **benchmark contamination** [20], i.e. spurious overfitting to specific numerical values or surface templates and ensures that observed performance is not driven by a memorization like sensitivity to encountering a math problem previously used in its training data.

For code-augmented experiments, we restrict representations to story, symbolic and word-equation forms to reduce compounding interface complexity.

Problem selection was intentionally constrained to preserve exactness and interpretability of evaluation. We excluded problems requiring logarithmic computation, non-square fractional exponents (e.g., $N^{1/3}$), or inherent irrational numbers such as $\pi$. Only square root operations yielding exact rational results were permitted. These exclusions avoid introducing approximation ambiguity, symbolic algebra confounds and tolerance-based grading, which would obscure whether observed differences arise from representation alone or from numeric approximation artifacts.

Ground truth answers are specified in exact form. Valid correct answers are integers, terminating decimals or reduced fractions. When an answer cannot be expressed as an integer or terminating decimal, it is represented as a pure fraction. This choice ensures deterministic auto grading, avoids arbitrary numeric tolerances and enforces a strict output contract across interaction conditions.

### 3.3 Interaction Conditions

We evaluate two interaction conditions that differ in how models are allowed to externalize reasoning.

In the ***baseline*** interaction condition, models are prompted to produce a direct final answer in text form. The system prompt enforces a strict output, that the response must consist only of the final answer, expressed as an integer, terminating decimal, reduced fraction, or a comma-separated list for multi answer problems. Explanations, intermediate reasoning, or extraneous text are not allowed. LLM outputs are parsed and normalized prior to scoring. Responses that violate the output contract (e.g., include explanations or malformed expressions) are recorded as ***format failures*** rather than discarded and are treated as incorrect but valid responses during analysis. Because baseline interactions are logged in a single pass, all returned model outputs, successful or malformed, contribute to an analyzable sample.

In the ***code-augmented*** executable reasoning interaction condition, models are required to output Python code bound by explicit delimiters. The code must compute the solution and assign the final value to a designated variable. No natural language output is permitted. Generated code is executed client-side in a restricted environment. Imports, file system access, network access, and system calls are disallowed. Only a small set of safe utilities is available, including exact rational arithmetic and square root operations when they yield exact results. Code execution is subject to strict timeouts. Failures in this interaction condition are explicitly categorized. ***Protocol failures*** occur when the model violates the output contract (e.g., missing delimiters or forbidden constructs). ***Execution failures*** occur when code raises runtime errors or exceeds time limits. These failure types are logged separately from incorrect but valid computations (***math failures***). We retain these failures because, from a user or system perspective, malformed outputs still consume time, cost, and cognitive effort, and therefore constitute meaningful interaction outcomes rather than noise to be filtered. To ensure comparability across tasks, code-augmented interactions use a completion-based execution policy where any transient infrastructure failures are retried and only completed executions are included in the analyzable sample.

### 3.4 Model Selection and Grouping Strategy

We evaluate 5 LLM models, organized by deployment and design role besides performance: Group A consists of frontier, reasoning heavy models (openai/gpt-5.2-chat and google/gemini-2.5-pro). Group B includes high quality general-purpose comparators (anthropic/claude-3.5-sonnet and x-ai/grok-4.1-fast) while Group C comprises open-weights baselines (meta-llama/llama-3.3-70b-instruct). This grouping follows prior

LLM evaluation work showing that reasoning behavior varies systematically with training scale, alignment objectives, and model accessibility( [21], [22]). Frontier proprietary models, general purpose commercial models, and open weights baselines exhibit distinct and well documented reasoning characteristics and failure modes( [23], [24], [25]). Organizing models along these deployment lines thus aligns with established benchmarking practice and enables analysis across qualitatively different model classes rather than a single accuracy ranking. ***All the 5 LLMs were included in the basic and code-augmented interaction conditions.***

### 3.5 Execution Framework

All model interactions are conducted through OpenRouter [26] which provides a unified API interface across multiple model providers. Using a single execution framework reduces provider specific implementation variance and enables consistent logging of usage metadata.

All experiments use fixed decoding settings, including deterministic decoding (temperature set to zero) and bounded maximum token counts. Task order is randomized with a fixed seed to mitigate ordering effects during long experimental runs. The setup includes retry and timeout logic and supports resuming partially completed experiments without duplicating completed calls.

### 3.6 Logging, Metrics, and Failure Categorization

For every model call, the system logs raw outputs, parsed and normalized answers, correctness labels, latency, token usage and estimated API cost. In the code-augmented interaction condition, code level features and execution outcomes are also recorded. Incorrect answers, format violations, protocol violations and execution errors are all retained and categorized. This design choice reflects the study's focus on LLMs as interactive systems (from a user and system perspective, bad outputs and execution failures incur real time, cost, and reliability penalties and therefore constitute meaningful outcomes).

The analysis collates outcomes across the representation variants, interaction conditions and model groups.

## 4 Results

### 4.1 Representation Robustness Under Constrained Interaction (RQ1)

We first compare the robustness of LLM performance to representationally equivalent formulations of the same mathematical problem. Fig. 1 shows accuracy aggregated across all models for each representation type under both interaction conditions.

For each model m and interaction condition c, we consider the subset $\mathcal{P}$ of problems for which all shared representations (story, symbolic, and wordeq) are available. For each representation r, accuracy is computed as the fraction of problems in $\mathcal{P}$ answered correctly by the model under representation r and condition c:

$$A_r(m,c) = \frac{1}{|\mathcal{P}|}\sum_{p\in\mathcal{P}} 1\left(\widehat{y_{m,c,r}}(p) = y(p)\right) \quad (1)$$

To obtain the aggregated accuracy $\overline{A_r}(c)$ shown in Fig. 1, we further average per-representation accuracy across the set of evaluated models $\mathcal{M}$:

$$\overline{A_r}(c) = \frac{1}{|\mathcal{M}|}\sum_{m\in\mathcal{M}} A_r(m,c) \quad (2)$$

Despite deterministic decoding and identical underlying mathematical structure, performance varies across representations as seen in Fig. 1. **This effect is present in both interaction conditions showing that representational sensitivity is not an artifact of sampling variability.**

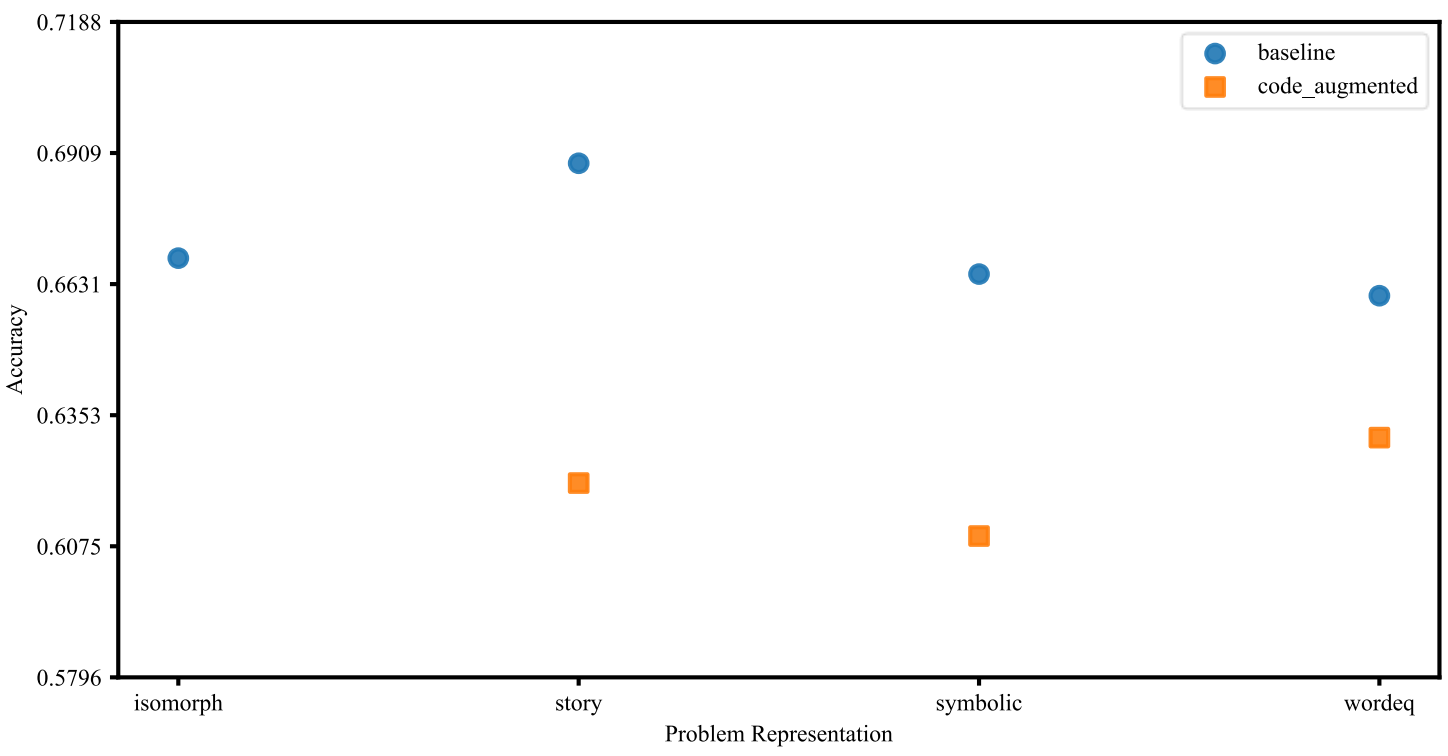

**Fig. 1.** Representation Sensitivity as aggregated accuracy $\overline{A_r}(c)$ across Interaction Conditions (Aggregated across models). Isomorph variant was applied only to baseline interaction condition. *For readability, the y axis is truncated and all accuracies lie in [0,1] and observed gaps are modest.*

Using the same per model, per representation accuracies $A_r(m,c)$, we define a model's ***max min representation gap G(m,c)*** under interaction condition $c$ as the range of accuracies across shared representations:

$$\text{G(m,c)}= \max_{r\in\{\text{story,symbolic,wordeq}\}} A_r(m,c) - \min_{r\in\{\text{story,symbolic,wordeq}\}} A_r(m,c) \quad (3)$$

Fig. 2 decomposes this effect at the model level by showing each model's representation gap, defined as the difference between its best and worst accuracy across shared representations (excluding isomorphic). While representation gaps are nontrivial for all models, their magnitude varies considerably. Some models exhibit relatively stable performance across representations, while others show pronounced sensitivity. Notably, some models (e.g., Meta Llama) exhibit increased representation gaps under executable reasoning, suggesting that code-based interaction can amplify representational specialization rather than uniformly stabilizing performance. ***Importantly, shifting from the*** ***baseline to the code-augmented interaction condition does not uniformly reduce these gaps.*** Instead, the interaction condition reshapes representational robustness in model-specific ways.

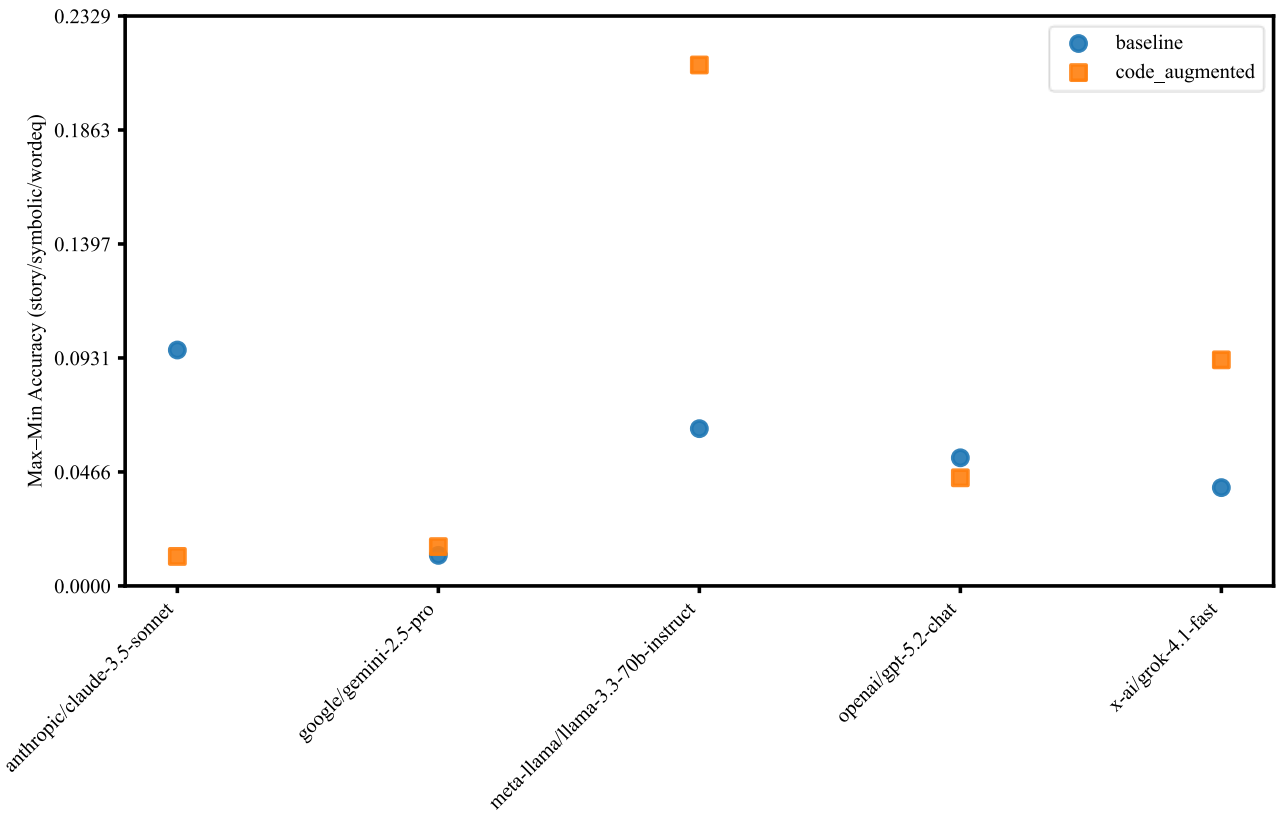

**Fig. 2.** Representation Gap G(m,c) by Model (Max Min across shared representations). Gap is computed as the difference between the maximum and minimum accuracy over story, symbolic, and word-equation variants.

To assess whether the observed representational differences between interaction conditions are statistically reliable, we further conducted representation-specific chi-square tests comparing baseline and code-augmented performance on the shared representations (story, symbolic, and word-equation). After applying a Bonferroni correction to control the family-wise error rate across three tests ($\alpha = 0.05/3 \approx 0.0167$), ***differences between interaction conditions were statistically significant for story problems ($\chi^2 = 13.14$, $p = 0.00029$) and symbolic problems ($\chi^2 = 8.44$, $p = 0.0037$), but not for word-equation problems ($\chi^2 = 2.63$, $p = 0.105$).***

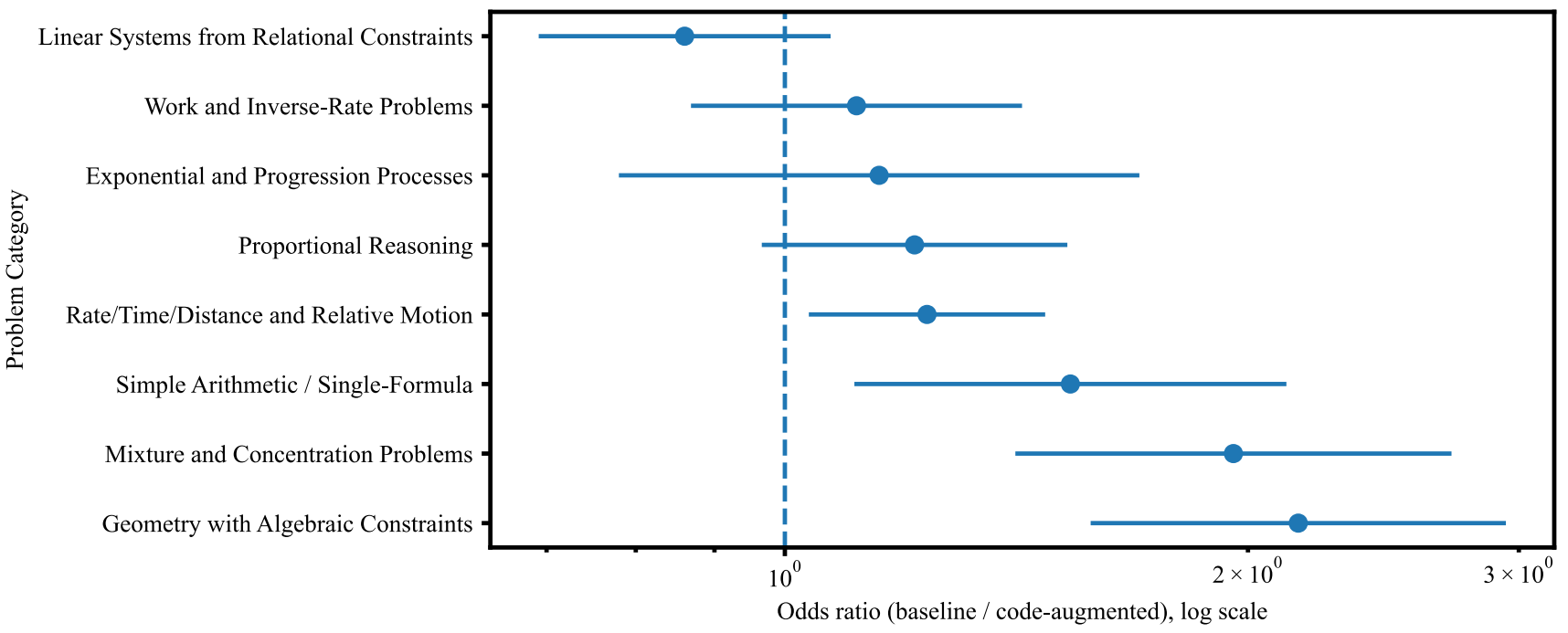

**Fig. 3.** Dataset content specific baseline/code differences across problem categories, showing how interaction condition effects vary by problem type. Points indicate the direction and magnitude of baseline versus code-augmented effects showing that executable reasoning reshapes representational sensitivity in a non-uniform manner rather than uniformly improving robustness.

However, effect sizes measured by Cramér's V(φ for 2×2 tables [27]) were small across all representations of story, symbolic, and word-equation ($V_{story} = 0.073$, $V_{symbolic} = 0.058$, $V_{wordeq} = 0.032$), indicating that ***even statistically detectable differences correspond to modest practical shifts in accuracy***. While representation level effect sizes are small, Fig. 3 visualizes these content specific effects by plotting the direction and magnitude of baseline-code differences across problem categories.

***Collectively, these results show that representational robustness is limited and uneven across models, and that constraining interaction through executable interfaces reshapes, rather than resolves, this sensitivity.***

### 4.2 Failure Structure and Interaction Layers (RQ2, RQ3)

We next analyze how interaction conditions alter not just accuracy but also the structure of failures. Fig. 4 presents failure composition aggregated across models for both interaction conditions. In the baseline condition, failures are divided into only two categories (incorrect or malformed responses, i.e. responses that did not abide by the system prompt). By contrast, the code-augmented condition allows for a layered failure structure separating protocol violations, execution failures, and mathematically incorrect computations.

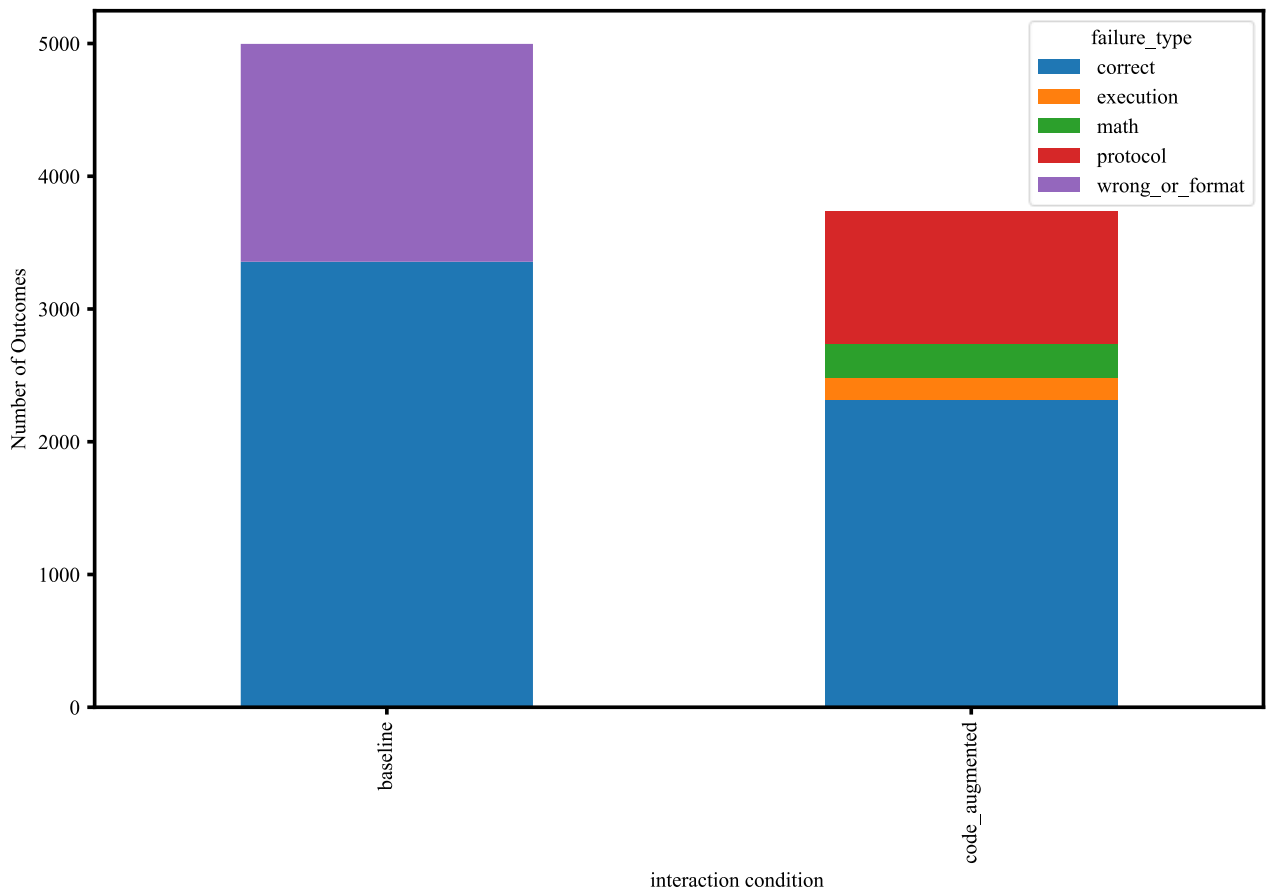

**Fig. 4.** Failure Structure by Interaction Condition (Aggregated across models)

Fig. 5 disaggregates these failures at the model level for the code-augmented condition. Models differ markedly in where they fail. Some frequently violate protocol constraints, others generate executable code that fails at runtime, and others execute successfully but compute incorrect results. These distinctions are invisible in baseline evaluation but become explicit when reasoning is externalized into Python code as executable artifacts.

Notably, in Fig. ***5 Gemini 2.5 Pro exhibits a markedly different failure profile under the executable reasoning condition***, with the majority of errors arising from protocol violations rather than execution or mathematical failures. This suggests that the model often produces correct or near-correct reasoning content that nonetheless fails to comply with the strict output contract required for executable interaction. This behavior is consistent with prior developer reports indicating that Gemini models are less amenable to fully suppressing natural language scaffolding ( [28], [29], [30]), showing that ***interaction condition constraints can expose model-specific mismatches between reasoning capability and interface compliance.***

Fig. 6 further characterizes this layered structure by relating math accuracy to execution reliability. Accuracy conditioned on successful execution does not fully capture system reliability since some models achieve high accuracy when execution succeeds but fail to execute reliably in the first place. This demonstrates a key interaction level distinction between being correct when working and being dependable as a system component.

These results collectively show that externalizing reasoning into executable code fundamentally changes failure behavior. ***From an HCI perspective, these layered failures are not equivalent: protocol violations, execution failures, and incorrect results impose qualitatively different burdens on users, even when they are collapsed into a single 'incorrect' label in benchmark evaluations.***

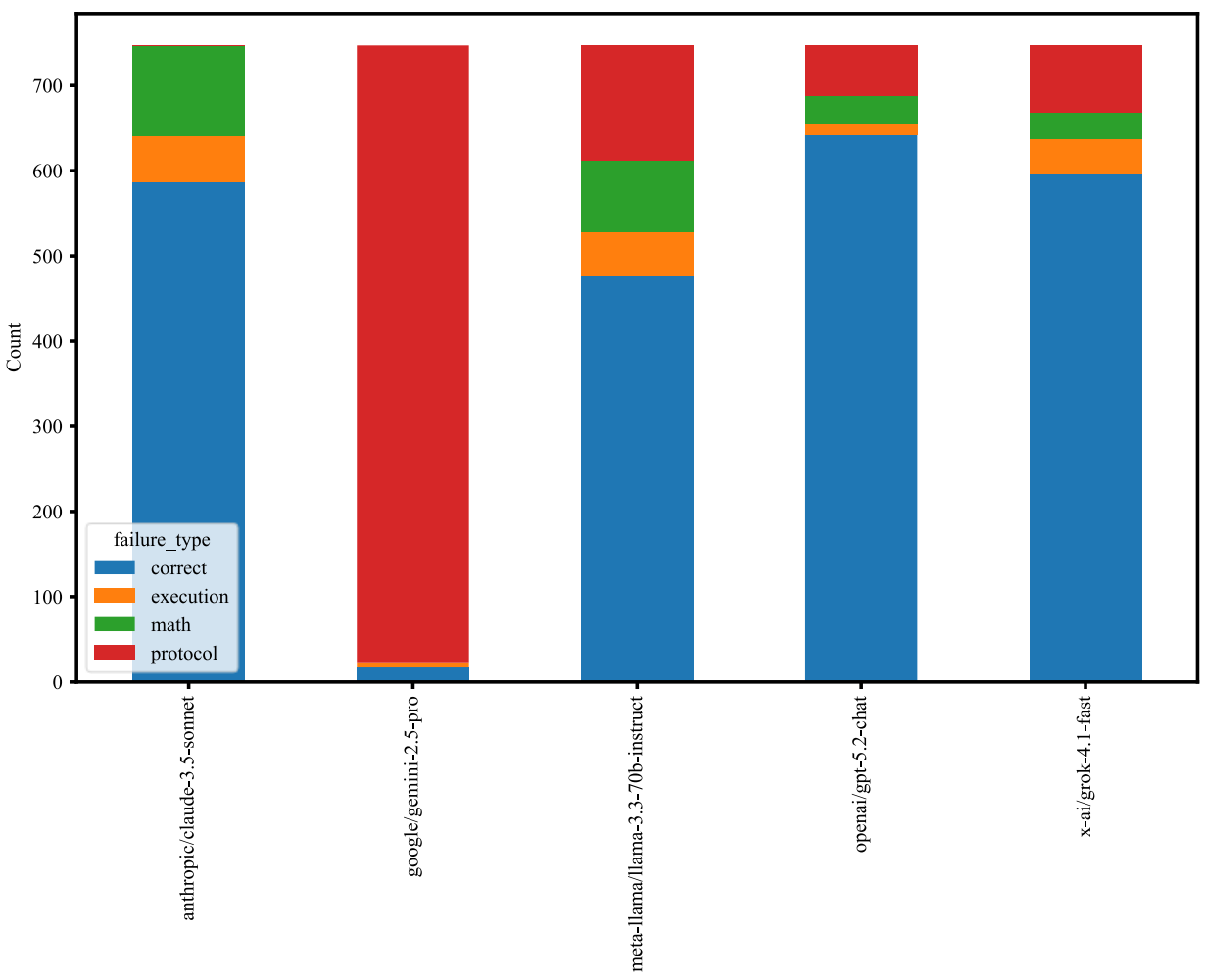

**Fig. 5.** Failure Structure by the code-augmented interaction condition (aggregated across models). Gemini 2.5 Pro notably exhibits a disproportionate number of protocol violations, indicating difficulty adhering to the strict code-only output contract rather than failures of execution or mathematical reasoning.

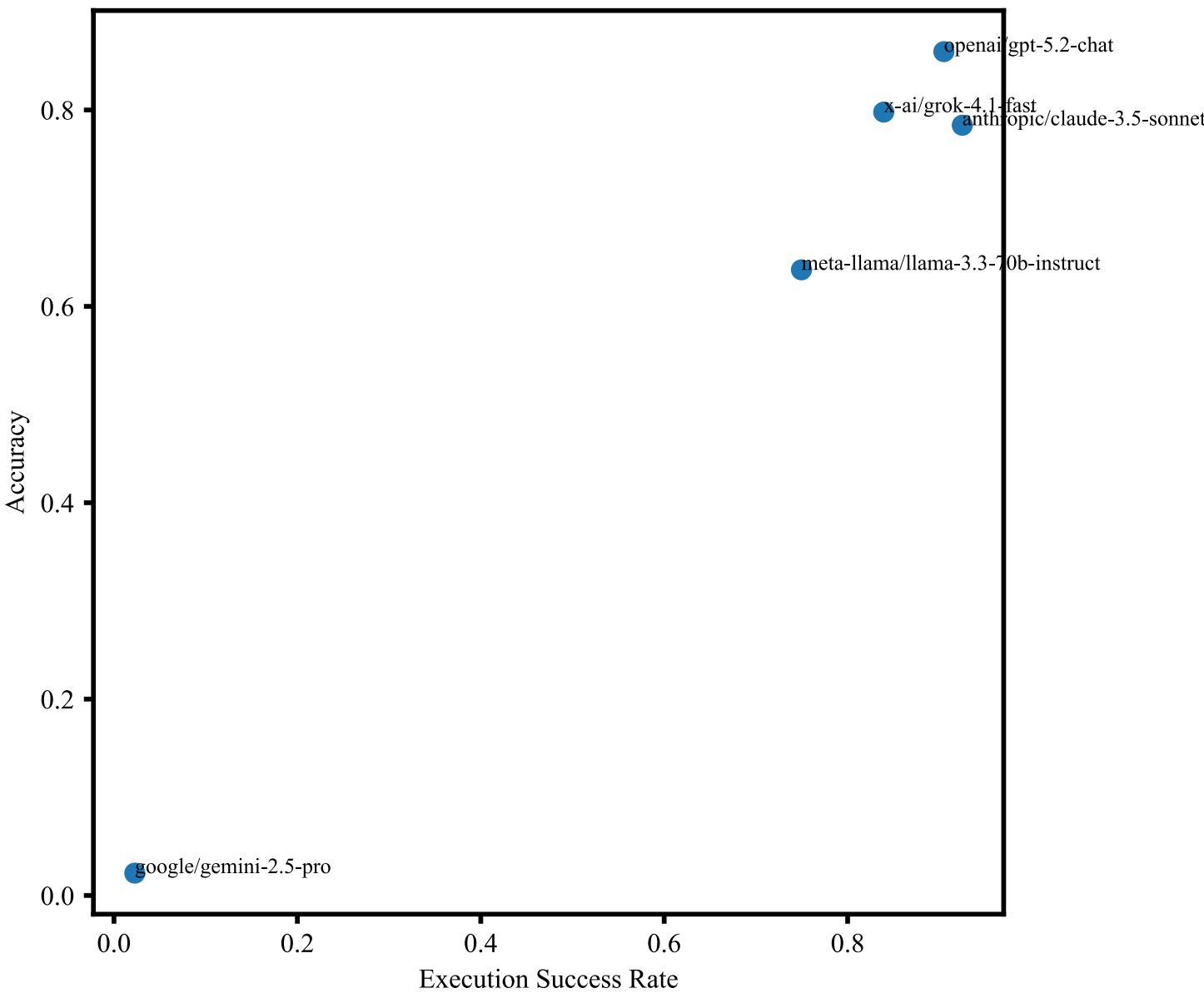

**Fig. 6.** Accuracy vs Execution Reliability (code-augmented only)

### 4.3 Cost and Latency as Interaction Consequences (RQ4)

Accuracy alone does not capture the user or system-level consequences of different interaction conditions. Fig. 7 shows median latency with one-sided p90 tails for each model under both interaction conditions. The code-augmented condition consistently increases both median latency and tail latency indicating not only slower responses but also greater variability in response time with respect to the baseline condition. This tail behavior is particularly relevant for interactive systems where occasional long delays can disproportionately affect user experience.

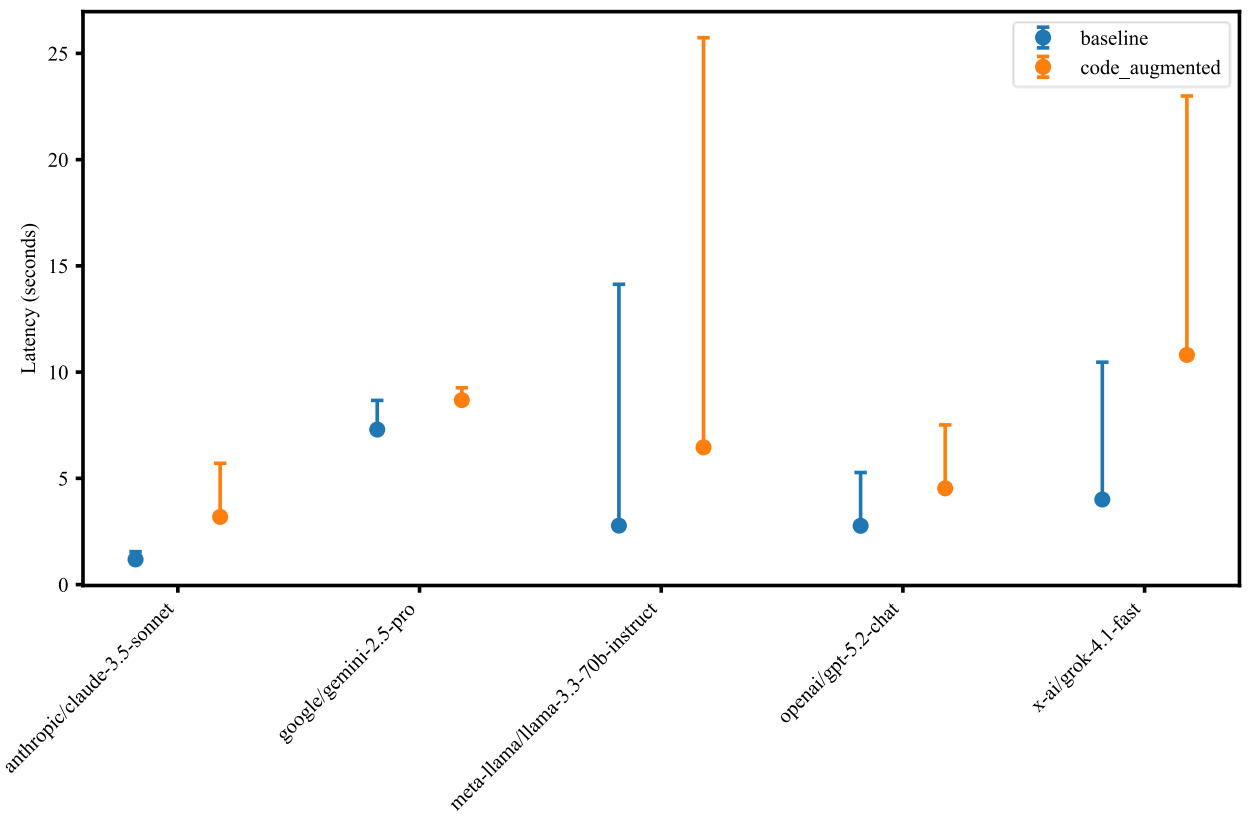

**Fig. 7.** Latency as Interaction Cost (model level for both interaction conditions)

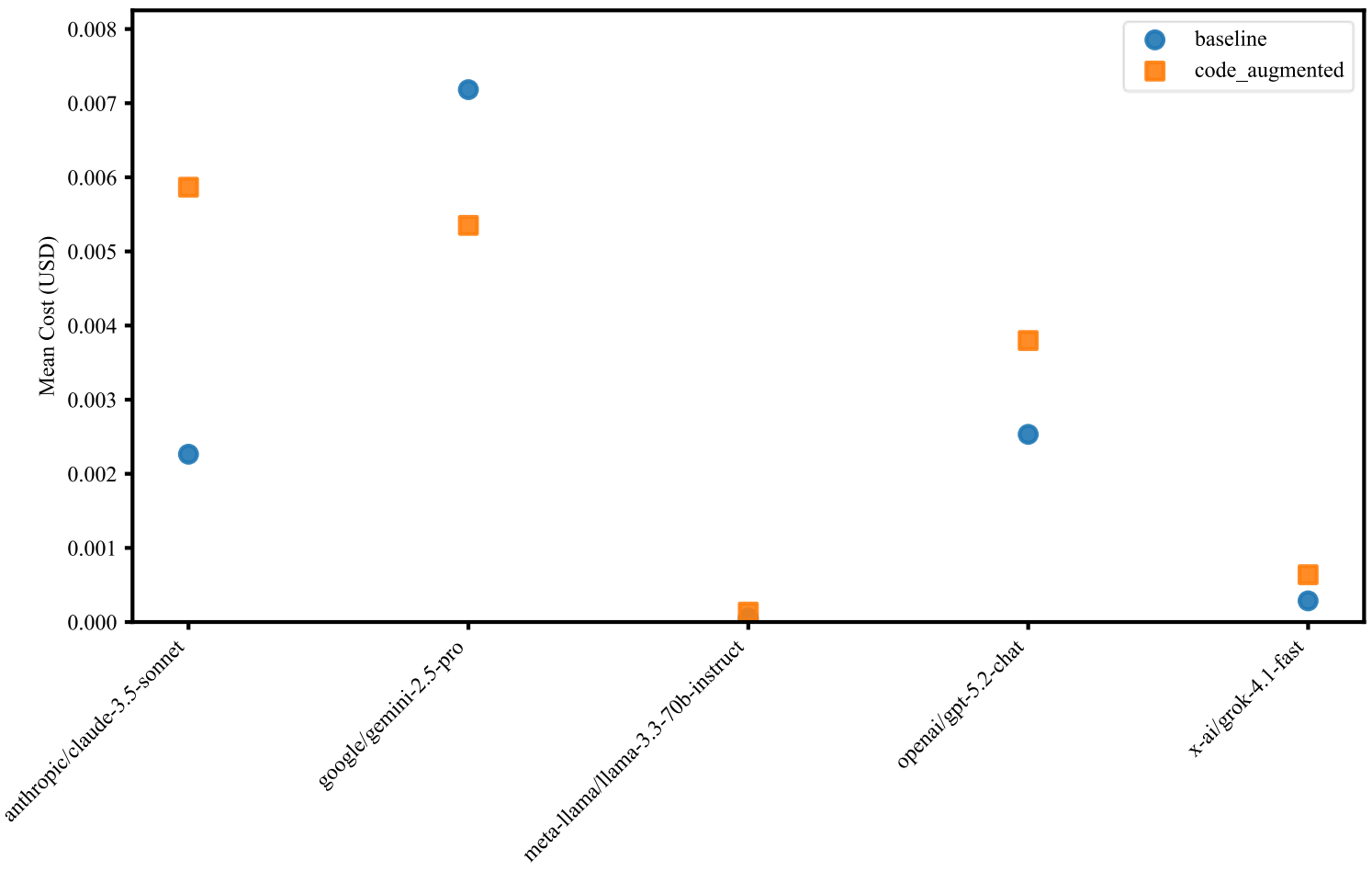

**Fig. 8.** Economic Cost per Correct Answer (model level for both interaction conditions)

Fig. 8 provides economic cost per correct answer. Costs typically increase under the code-augmented condition, reflecting the cumulative expense of failed executions, retries and longer responses. Gemini 2.5 Pro constitutes a notable outlier, suggesting that for models with strong latent reasoning but brittle interface compliance, executable constraints can reduce per-success cost despite increasing overall failure rates. ***Models that appear competitive under accuracy metrics alone can be substantially less efficient when evaluated in terms of cost per successful outcome.***

These results highlight that reasoning scaffolds introduce hidden interaction costs, making models that appear strong under accuracy only evaluations potentially unsuitable for latency-sensitive or cost-constrained settings.

### 4.4 Tradeoffs and Interaction Condition Level Shifts (RQ4 Synthesis)

We also collate accuracy, latency, and reliability into a single interaction level view. Fig. 9 superimposes baseline and code-augmented results in a shared latency accuracy space. Each model occupies distinct regions under the two interaction conditions, making visible the cost of mediation through executable interfaces. It is shown that no model dominates across both accuracy and latency and improvements in one dimension are often accompanied by regressions in another.

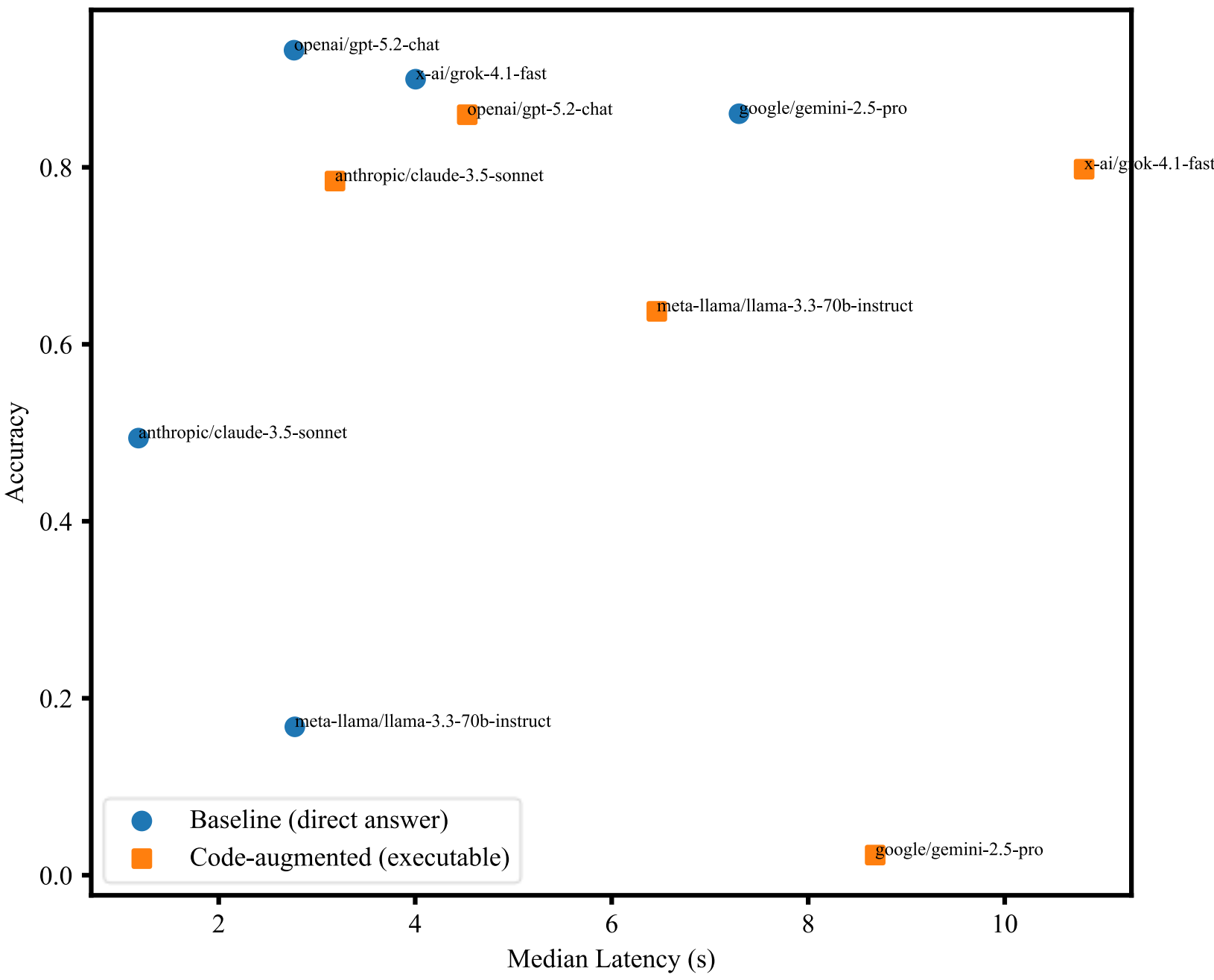

**Fig. 9.** Latency Accuracy Tradeoffs (model level for code-augmented conditions)

Taken together, these tradeoffs show that interaction conditions do not simply improve or degrade model performance. Instead, they alter system behavior, shifting where and how costs are incurred. Evaluating LLMs solely on benchmark accuracy obscures these shifts and risks overstating reasoning stability.

Across all four RQs, the results demonstrate that LLM performance is not invariant to representation or interaction design. Accuracy, failure structure, latency, and cost are tightly coupled, and meaningful evaluation requires treating LLMs as interactive systems rather than isolated problem solvers.

## 5 Implications for the Design and Evaluation of AI Assisted Problem Solving Systems

Our results suggest that many commonly reported measures of LLM reasoning performance conflate problem difficulty, surface representation, and interaction design. Rather than treating these effects as model quirks, we interpret them as signals of misalignment between how LLMs are evaluated and how they are deployed as interactive systems. This section discusses implications for benchmark construction, system design, and evaluation practice.

### 5.1 Implications for Benchmark and Dataset Design

Standard math benchmarks often treat representationally equivalent problem formulations as interchangeable, implicitly assuming that variation in surface form does not materially affect underlying reasoning by LLMs [31]. Our findings challenge this assumption. Even under deterministic decoding, performance varies substantially across representations of the same mathematical structure and these variations persist across interaction conditions.

From an evaluation perspective, aggregating accuracy across representations can obscure systematic brittleness. ***High benchmark scores may therefore reflect sensitivity to favored surface forms rather than stable representation-invariant ability***. ***For benchmark designers, this suggests a shift away from single form problem instances toward controlled families of representational variants.*** Such datasets make it possible to distinguish robustness from overfitting to the presentation style.

***More importantly, this implication is not limited to mathematics***. Any domain that admits multiple surface realizations of the same underlying structure, such as programming tasks, word problems, legal reasoning, or instructional dialogue faces similar potential risks when representation is treated as incidental.

### 5.2 Implications for Interactive System Design

Introducing constrained executable reasoning through a combination of system prompts and tool use such as code augmentation does not eliminate failures as it redistributes them across interaction layers. In the code-augmented interaction condition, errors emerge at the level of protocol adherence, execution, and mathematical correctness. These failure types are qualitatively different from the user's perspective even when they are equivalent under a binary correct/incorrect metric.

For system designers, this suggests that adding structure to the model outputs should be understood as a tradeoff rather than an unambiguous improvement. While constraints can increase transparency and debuggability, but they also introduce new points of breakdown. Designing AI assisted math problem solving systems therefore requires anticipating how failures migrate across representation layers rather than assuming they will disappear. ***This perspective aligns with treating LLMs as components within socio-technical systems where interaction contracts, execution environments and error handling mechanisms shape user experience as much as raw model capability.***

### 5.3 Implications for Reliability Aware Evaluation

Our results indicate that correctness alone is an incomplete measure of system quality. ***Failures that consume time, incur financial costs or require repeated attempts impose real penalties on users and developers even when the final answer is correct.*** Tail latency, execution reliability and cost per successful outcome can therefore emerge as first-class properties of system behavior.

Evaluation frameworks that ignore these factors risk overstating practical performance. For example, a model that is highly accurate conditional on successful execution but frequently fails to execute may be unsuitable for interactive use despite strong benchmark scores. Reliability aware evaluation therefore requires tracking not only what answers models produce but also how consistently and at what cost they produce them.

## 6 Future Work

Future work could extend this study through several dimensions. Investigating larger and more diverse problem families would enable finer-grained analysis of representational robustness beyond controlled mathematical probes including domains such as programming, planning and instructional dialogue. Also, incorporating richer interaction conditions (e.g. such as multiturn repair, user in the loop, tool assisted fallback strategies) could reveal how failure layers evolve over longer interaction horizons. Additionally, longitudinal evaluations that track adaptation, caching, or prompt drift over time would help distinguish transient fragility from persistent system level brittleness. Integrating human subject studies into a future study could also connect observed failure structures to user trust, frustration, and task success, closing the loop between system behavior and human experience.

An additional direction for future work is longitudinal evaluation across successive generations of models from the same provider. Applying the same controlled representational probes to model descendants would allow for tracking how representational robustness and failure structure evolve over time. Even under black box model access constraints, such comparisons can reveal whether improvements in benchmark accuracy correspond to increased invariance across representations or merely shifts in which surface forms are favored.

**Disclosure of Interests.** The authors have no competing interests to declare that are relevant to the content of this article.